\documentclass[10pt, conference, compsocconf]{IEEEtran}
\ifCLASSINFOpdf
\else
\fi
\usepackage[cmex10]{amsmath}
\usepackage{amssymb}
\usepackage{dsfont}

\usepackage{makecell}
\usepackage{algorithm}
\usepackage{algcompatible}
\usepackage{algpseudocode}

\usepackage{graphicx}
\usepackage{multirow}
\usepackage{svg}

\makeatletter
\newcommand{\thickhline}{%
    \noalign {\ifnum 0=`}\fi \hrule height 1pt
    \futurelet \reserved@a \@xhline
}
\newcolumntype{"}{@{\hskip\tabcolsep\vrule width 1pt\hskip\tabcolsep}}
\makeatother

\begin{document}
%
\title{Multi Player Tracking in Ice Hockey with Homographic Projections}


\author{\IEEEauthorblockN{Harish Prakash, Jia Cheng Shang, Ken M. Nsiempba, Yuhao Chen, David A. Clausi, John S. Zelek}
\IEEEauthorblockA{Vision and Image Processing Group \\ 
University of Waterloo, Ontario, Canada\\
\{harish.prakash, jcshang, kmnsiemp, yuhao.chen1, dclausi, jzelek\} @ uwaterloo.ca
} }


%


\maketitle

\begin{figure*}
\centering
\includegraphics[width=17.7cm]{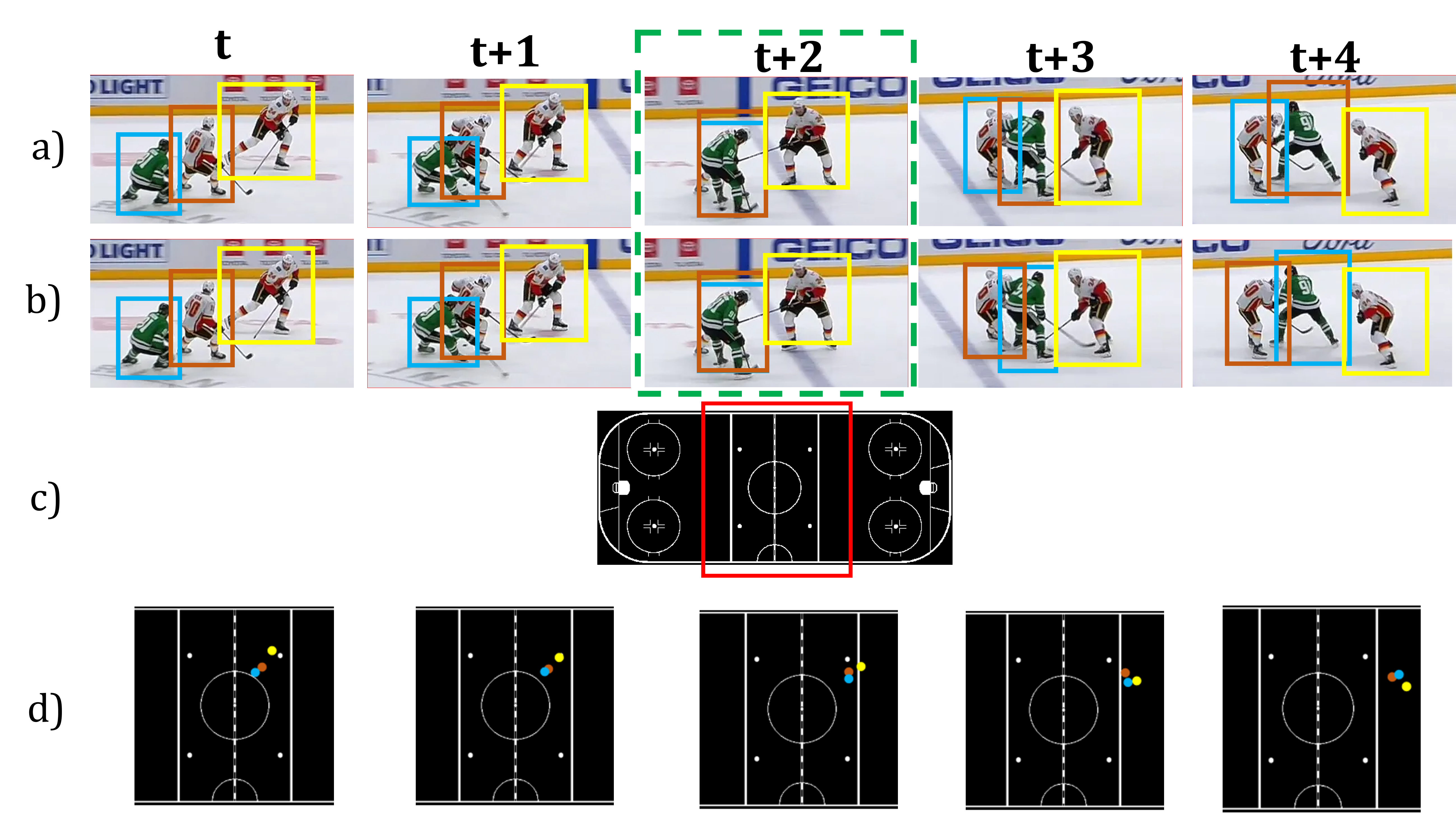}
\caption{Other trackers vs Our approach. a) During a significant occlusion scenario $(t+2)$, other trackers lead to an ID switch error; b) Our method consistently tracks players before and after occlusion; c) Overhead rink template used for homography projection; d) Player footpoint coordinates mapped to the overhead template. At $(t+2)$, there is a clear distinction between overlapping players from the top view. This information aids our tracker in maintaining player tracklets. 
}
\end{figure*}

\begin{abstract}
Multi Object Tracking (MOT) in ice hockey pursues the combined task of localizing and associating players across a given sequence to maintain their identities. Tracking players from monocular broadcast feeds is an important computer vision problem offering various downstream analytics and enhanced viewership experience. However, existing trackers encounter significant difficulties in dealing with occlusions, blurs, and agile player movements prevalent in telecast feeds. In this work, we propose a novel tracking approach by formulating MOT as a bipartite graph matching problem infused with homography. 
We disentangle the positional representations of occluded and overlapping players in broadcast view, by mapping their foot keypoints to an overhead rink template, and encode these projected positions into the graph network. This ensures reliable spatial context for consistent player tracking and unfragmented tracklet prediction. Our results show considerable improvements in both the \emph{IDsw} and \emph{IDF1} metrics on the two available broadcast ice hockey datasets.


\end{abstract}


%
\IEEEpeerreviewmaketitle

\section{Introduction}
Multi-Object Tracking (MOT) is an important computer vision problem subsuming several tasks such as object detection, localization, re-identification and association across a temporal sequence. It is a highly studied and established problem due to its plethora of applications in robotics, autonomous vehicles, industrial automation, surveillance, and sports. While most existing MOT approaches focus exclusively on tracking crowded pedestrians \cite{mot2015, mot16, mot20}, group dancing \cite{dancetrack}, and autonomous driving \cite{kitti-autonomous-vehicles}, sports tracking is a pivotal task in vision due to its numerous subsequent  applications in game analytics \& statistics, strategic planning, player evaluation, injury prevention, and crucial game decisions. Player tracking helps save several manual labor hours and human efforts by automating game understanding and player performance assessments. With the advent of deep networks\cite{alexnet}, several important strides have been taken to track various sports including soccer \cite{soccertrackingsurvey}, handball \cite{handball-tracking}, basketball \cite{basketball1, basketball2, basketball3}, and volleyball \cite{volleyball1}, with public MOT datasets \cite{sportsmot, soccernet} to support principled evaluations. 

Unlike all the aforementioned sports, ice hockey poses unique challenges to tracking due to its highly \emph{physical} and \emph{fast-paced} nature. Specifically there are three major challenges that exists: \textbf{(i)} the significant occlusion between two or more players at a given instant within the field-of-view (FoV); \textbf{(ii)} the non-linear player dynamics due to unpredictable player motion, and; \textbf{(iii)} blurs and reduced visibility of players due to frequent camera motion. When faced with these issues, tracking in monocular view increases identity swaps and tracklet fragmentations. Early attempts at tracking ice hockey players were pursued using an ensemble of handcrafted methods. Okuma et al. \cite{okuma} use Adaboost \cite{adaboost} detection with mixed particle filters \cite{mixture-particle-filter} for tracking players from television videos. Cai et al. \cite{cai} improve upon \cite{okuma} by utilizing the mean-shift algorithm to stabilize player trajectories, and use rink coordinates (homography) for the particle-filter. However, mixed particle filters are susceptible to identity switches/losses during mutual occlusions, background changes, blurs, and lighting effects, which are often found in hockey. Further, there exists no quantitative evaluation in the above works to show the efficacy of their models. 

Recent approaches using deep networks have shown significant improvements in the accuracy of tracking hockey players from broadcast feeds. Vats et al. \cite{vatsevaluation} present a comparison between five different tracking models, fine-tuned on broadcast ice hockey clips and obtain state-of-the-art results using graphs with message passing networks (MPN) \cite{motneuralsolver}. They subsequently use this method to generate tracklets (sequence of player tracks) for player identification \& team recognition \cite{vatstrackingandid}. As far as we know, this is the only existing benchmark for MOT in ice hockey. However, there exists two major limitations in their approach: first, their model encodes the \emph{bounding box} attributes of players as graphical edge embeddings which leads to identity swaps and tracklet fragmentations. This is because, when there exists heavy occlusion between players as usually observed in broadcast hockey feeds, their bounding boxes overlap significantly in the monocular view (Intersection-over-Union (IoU) $\uparrow$), 
causing either misassociation of players (identity switch) or a missed connection (lost tracklet) with previous tracks. Second, their node embeddings are encoded solely based on player appearance features, which is ambiguous in hockey due to the fully-covered bulk gear worn by players, similarly colored team jerseys, blur \& lighting effects. With these present setbacks, we ask the question: \emph{"Given only a monocular broadcast feed, is it possible to declutter occluded players and track their movements with high fidelity?"}. Our results show that the answer is \textbf{Yes}. 

In this work, we formulate MOT as a link prediction problem in the graph domain since graphical networks offer a natural way of representing players and their relationships. The novelty of our approach lies in coupling this with homography to obtain reliable spatial/positional cues for ice hockey players. Specifically, we map every frame from broadcast view to an overhead view (rink template) using an off-the-shelf homography estimator \cite{jason-homography}, to obtain the projected footpoints of players in rink coordinates. This yields a \emph{birds-eye} effect, reducing the positional ambiguities due to overlapping players and perspective projection. We encode these projected positions along with player re-identification features as graph embeddings, and propagate information through the graph using the message passing network (MPN) \cite{mpn, mpn-quantumchem} framework. Our model follows the popular \emph{tracking-by-detection} paradigm and we show evaluations with both ground-truth detections and off-the-shelf detector outputs \cite{faster-rcnn}, in coherence with the current state-of-the-art (SOTA) benchmark \cite{vatsevaluation}. Through this approach, our model is able to utilize the overhead positional information of players as additional cues to track with consistency during occlusions, blurs, and dynamic player movements.

Our contributions can be summarized as follows: 
\begin{itemize}
\item We design a simple, yet effective spatio-temporal graphical network and adopt the MPN framework to track players from broadcast feeds consistently;  
\item We propose a novel approach based on homography to provide additional positional information to the graph network during occlusions, blurs and non-linear movements, and; 
\item We show significant improvements in the \emph{IDsw} and \emph{IDF1} scores on two available broadcast ice hockey datasets.
\end{itemize}

\section{Related Works}

\subsection{MOT for Pedestrians} 
Most existing MOT methods \cite{motneuralsolver, smiletrack, ucmctrack, bytetrack, tracktor, transmot, SORT, deepSORT, fairmot} focus exclusively on tracking pedestrians in crowded scenes \cite{mot2015, mot16, mot20}. Initial approaches combine Kalman filters \cite{kalmanfilters} and Hungarian method \cite{hungarianmethod} for next state motion prediction and object associations respectively. 
Consequent approaches adopted the two-stage \emph{tracking-by-detection (TBD)} paradigm where: first, all objects present in a sequence are detected; and next, their association features are extracted to link similar objects across frames. Simple Online and Real-time tracking (SORT) \cite{SORT} establish a TBD baseline for MOT, and argue that SOTA associations can only be obtained with traditional methods \cite{kalmanfilters, hungarianmethod}. DeepSORT \cite{deepSORT} answer SORT's \cite{SORT} argument, by embedding appearance features using a ReID network for association, showing lesser identity switches and better tracking results on the MOT16 \cite{mot16} challenge benchmark. FairMOT \cite{fairmot} combines the detection and ReID steps for Joint Detection and Tracking (JDT) using an anchor-free detector \cite{centernet}. Tracktor \cite{tracktor} frames tracking as a bounding box regression problem, by converting a detector \cite{faster-rcnn} into a tracker. But, a major limitation with all these methods is their inability to tackle crowded scenes and declutter occluded pedestrians. ByteTrack \cite{bytetrack} tries to handle occlusion by associating low-confidence targets too, to retrieve true objects, but fails during significant crowded (overlapping) situations where the object is heavily obstructed. Sparsetrack \cite{sparsetrack} tries to handle occlusion by decomposing dense/crowded pedestrian scenes into sparse subsets using pseudo-depth map estimation, but incur high computational costs and significant processing requirements.

\subsection{MOT for Sports}
Initial methods in sports utilize Kalman filtering \cite{iwase-soccer, ming-soccer} and particle filters \cite{okuma, cai} for tracking, but were unable to preserve identities due to their linear motion model assumptions. Nillius et al. \cite{bayesian-soccer} encode player trajectories into track graphs and use a bayesian framework to predict the most likely configuration of player paths. Figueroa et al. \cite{initial-graph-2004} track players with multi-cameras by encoding their segmentation blobs as nodes and their relative distances as edges. 
With YOLOv2 \cite{Yolo} for detection, Acuna et al. \cite{basketball-SORT} track basketball players using SORT \cite{SORT}, while Theagarajan et al. \cite{theagarajan} track soccer players using DeepSORT \cite{deepSORT}. However, naively extending methods originally designed for pedestrian tracking to sports is a non-trivial task, due to the domain-specific challenges in modeling arbitrary player movements, occlusions, fast camera motions, and perspective projection errors. 

\subsection{Tracking with Graphs}
Graph-based formulations offer a flexible way to model target movements, interactions and their relative features. Wang et al. \cite{jdegraphnetwork} propose a graph-based MOT framework for joint detection and tracking. MOT neural solver \cite{motneuralsolver} exploits the network flow formulation of MOT to define a message passing network for data association. Vats et al. \cite{vatsevaluation} fine-tune the neural solver architecture on the broadcast ice hockey dataset to create the first tracking benchmark for hockey.
Luna et al. \cite{graph-mcmot} extend the graph domain to multi-camera tracking, and ReST \cite{ReST} builds on top with a spatio-temporal network for online-tracking. Both these algorithms use the intrinsic camera parameters to map multiple camera views onto a common ground plane for additional positional cues. But, in our case, we cannot infer camera parameters from broadcast feeds directly. Therefore, we utilize an off-the-shelf homography model \cite{jason-homography} trained on a top-view rink template, to map each broadcast frame to the overhead rink coordinates and obtain homographic footpoint coordinates for the graph MPN.

\begin{figure*}
\label{proposed-workflow}
\includegraphics[width=17.7cm]{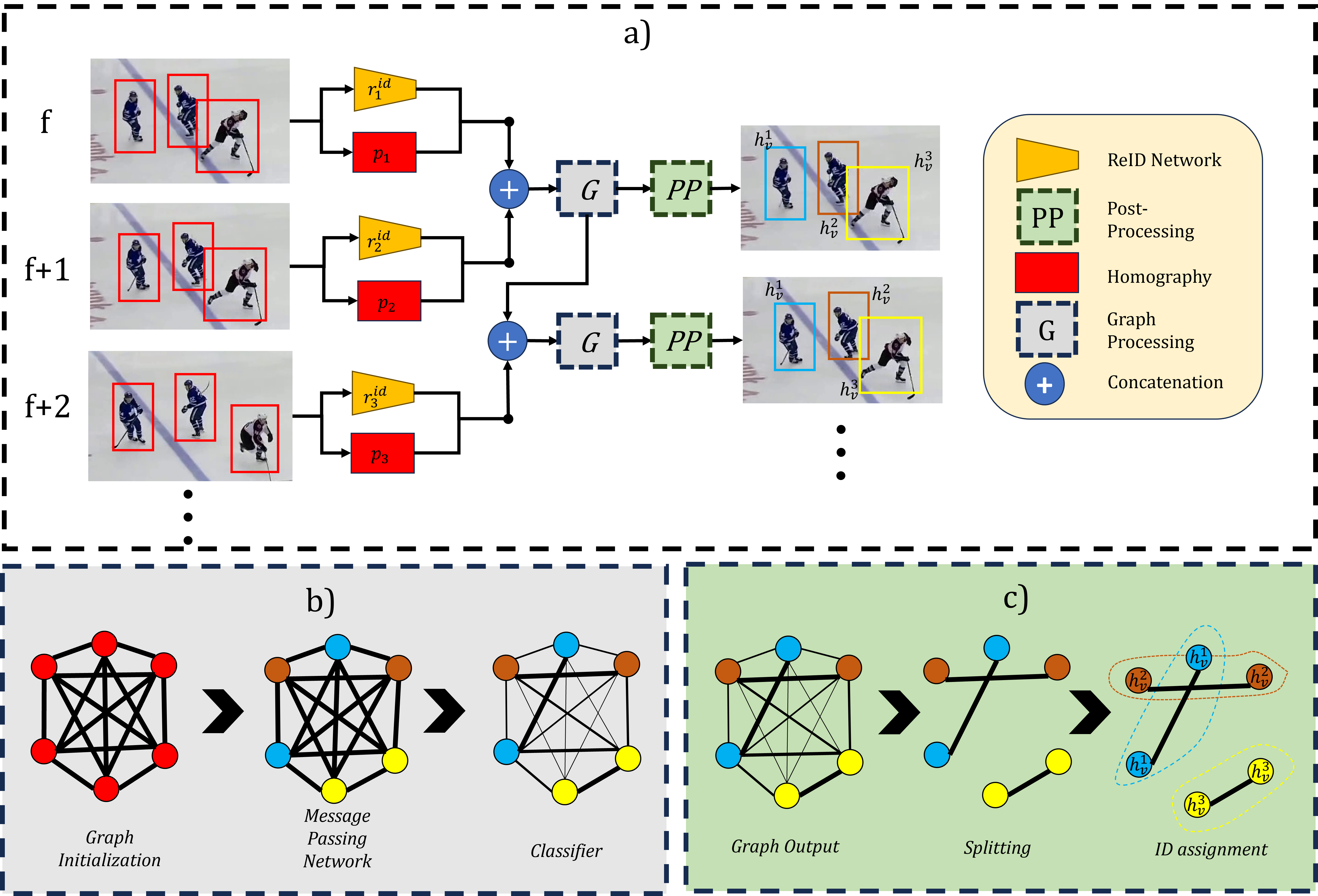}
\caption{Proposed Approach. a) The general pipeline of our spatio-temporal graph. b) $G$ denotes the three stages of Graph Initialization, MPN, and Classification. c) $PP$ denotes the Post-Processing stage where we Prune, solve graph violations, and assign player IDs}
\end{figure*}

\section{Proposed Method}

Given an ice hockey broadcast feed, the objective of our work is to track multiple players consistently despite prevalent occlusions, blurs, and arbitrary player movements. We leverage Graphical Neural Networks (GNN) for this task (due to their efficiency in modeling relationships) to build a spatio-temporal graph and utilize homography for reliable positional information. Specifically, we project player foot keypoints onto a common overhead rink template and encode these projections along with ReID embeddings \cite{omni-reid} as node features and their relative distances as edge features. Since we cannot obtain camera parameters from broadcast feeds directly, we use an off-the-shelf homography estimation model \cite{jason-homography} specifically designed for ice hockey, to estimate the homography transformation matrix, $H \in \mathbb{R}^{3 \times 3}$. We utilize the message passing network (MPN) framework \cite{mpn, mpn-quantumchem} to propagate player features across the entire graph ${G}$, and update the node and edge embeddings at each message passing step. This helps our model reason globally over the entire sequence for predicting player trajectories. We frame the association between two consecutive frames ${f_i}$ and ${f_j}$ ${s.t \ j > i}$, as a bipartite graph problem, and use a binary classifier with a sigmoid final layer to output association probabilities. During inference, we post-process each graph by pruning to remove low-confidence associations and solve many-to-one violations. Finally, we assign tracklet IDs to nodes wherein, if the node has a prior connection, it inherits the same ID or gets assigned a new ID otherwise. (Ref Fig. \ref{proposed-workflow})

\subsection{Problem Formulation}

Consider a broadcast hockey sequence (frames) $S_t = \{F_i \mid i = 1, 2, \ldots, n\}, \ \text{where \ } n = \frac{t}{\Delta t}$; $t$ = total duration of the sequence, and $\Delta t$ = duration per frame. For each frame $F_i$, assuming that there exists at least one player to track, we have $P_j^i$ players, where $j >= 1$. For each player $P$, the ground truth annotation includes: $$ \{ f^{id}, t^{id}, x, y, wd, ht, c, x^{proj}, y^{proj} \} $$ where $f^{id}$ = frame ID, $t^{id}$ = player ID (only used during training), $\{x,y,wd,ht\}$ = bounding box coordinates, $c$ = annotation confidence score, and $(x^{proj}, y^{proj})$ = homography coordinates for player footpoints. As per the \emph{tracking-by-detection} paradigm, player annotations are detections from an off-the-shelf detector for the inference stage. In our case, we show both evaluations: first, directly evaluating on the annotated ground truth which gives the best picture of our tracking performance (Ref. Section \ref{Evaluation}), and second, inference and evaluation on the detected output using F-RCNN \cite{faster-rcnn} following the current SOTA \cite{vatsevaluation} benchmark, for a fair comparison. 

\vspace{3pt}
We formulate multi-object tracking (MOT) as a bipartite graph matching problem. We deterministically create one graph per frame in the given sequence. Each node in this graph points to one player in that frame and the edges correspond to their relationships with neighboring nodes in other frames. Let us consider the undirected graph $G_t = (V_t,E_t)$ with bidirectional connections, where $V_t$ denotes the vertex set and $E_t$ denotes the edge set at time $t$. $NodeFeatures(.)$ represents the concatenation of appearance features and homographic projections of players at time $t$, and $E_t = V_i \times V_j$ represents a one-one mapping between vertex sets $v_i$ and $v_j \ s.t \ i \neq j$, meaning that the mapping is not within players in the same frame. 

\vspace{5pt}
\noindent \textbf{Node Formulation}
 Each node $v_i \in V_t$ represents one unique player $i$ found in $F_t$ and contains: frame ID (timestamp) $f^{id}_i \in \mathbb{R}^1$, player ID $t^{id}_i \in \mathbb{R}^1$ (ground-truth, only used for training), bounding box coordinates $b_i = \{x_i,y_i,wd_i,ht_i\} \in \mathbb{R}^4$, ReID appearance features $r^{id}_i \in \mathbb{R}^{512}$ and homographic projection coordinates $p_i = \{ x^{proj}_i,y^{proj}_i \} \in \mathbb{R}^2$.

\vspace{2pt}
The ReID features are generated for each node $i$ via an off-the-shelf Re-Identification network \cite{omni-reid}, described by: 
\begin{equation}
    \label{reid-equation}
    r^{id}_i = ReID(b_i \mid _{crop})
\end{equation}
where $b_i \mid _{crop}$ denotes the cropped bounding box for the $i^{th}$ player. The homographic projections are estimated by projecting the bottom-mid point ($\sim$foot keypoint) of the bounding box from the monocular broadcast view. The left footpoint $f_l = (x_i + \frac{wd_i}{2})$ and the right footpoint $f_r = (y_i + ht_i)$ for player $i$ are projected as:

\begin{equation}
\label{nodeformulation-homography}
    p_i = H_i(f_l, f_r)
\end{equation}

with $H_i$ being the $3\times3$ Homography matrix (Ref. Eq. \ref{homgraphy-matrix-eqn})

\vspace{5pt}
\noindent \textbf{Edge Formulation}  Each edge, $e_{ij} \in E_t$ is represented as the interconnection between two players from two distinct frames. It is encoded as the concatenation of relative appearance $\Delta r^{id}_{ij}$ and positional $\Delta p_{ij}$ features between the pair of nodes $v_i$ and $v_j$, where $e_{ij} =  v_i \times v_j, i \neq j$. This can be represented as: 

\begin{equation}
\label{edgeformulation-relative-reid}
    \Delta r^{id}_{ij} = [\lVert r^{id}_i - r^{id}_j\rVert_1,cosine\_similarity(r^{id}_i, r^{id}_j)]
\end{equation}

\begin{equation}
\label{edgeformulation-relative-homo}
    \Delta p_{ij} = [\lVert p_i - p_j\rVert_1, \lVert p_i - p_j\rVert_2]
\end{equation}

[·,·] denotes the concatenation of Eucledian distance \& Cosine Similarity for $\Delta r^{id}_{ij}$,  and Eucledian \& Manhattan distance for $\Delta p_{ij}$. This consideration is inspired from \cite{graph-mcmot} to obtain higher-dimensional distinctive features. 

\subsection{Homographic Projection}
Due to the monocular nature of broadcast ice hockey sequences, there exists high levels of player occlusion and dynamic camera movement effects (blurs, pans, tilts, zooms). This limits the scope of tracking players consistently when they're completely obscured or remain hidden, even if for very short intervals. To provide the tracker with reliable positional cues in such scenarios, we propose an approach using homography to warp player positions from the broadcast video feed onto an common overhead rink template. This helps reduce the variance in frames present across the sequence due to camera motion, and provides a pseudo \emph{top-view} tracking effect to disentangle overlapping players. 

At any given frame $F_t$, a player's $P_i$ foot keypoint coordinates represents their exact point of contact with the ice, which when projected to the overhead rink plane provides additional positional cues for uncluttered tracking. For the player $P$ with footpoints $(P_{f_x}, P_{f_y})$ in broadcast view: 

\begin{equation}
\label{homgraphy-matrix-eqn}
    p_i = 
s 
\begin{bmatrix}
    P_{x'} \\    
    P_{y'} \\ 
    1
\end{bmatrix}
= 
H
\begin{bmatrix}
    P_{f_x} \\    
    P_{f_y} \\ 
    1
\end{bmatrix}
=
\begin{bmatrix}
    h_{11} & h_{12} & h_{13} \\
    h_{21} & h_{22} & h_{23} \\
    h_{31} & h_{32} & 1 \\
\end{bmatrix}
\begin{bmatrix}
    P_{f_x} \\    
    P_{f_y} \\ 
    1
\end{bmatrix}
\end{equation}

where $p_i$ = homographic projection of the $i^{th}$ node,  $(P_{x'}, P_{y'})$ = projected homogenous player footpoints in the overhead view, $s$ = scale factor, and $H = 3 \times 3$ homography matrix.

But this is a non-trivial task, since the camera parameters for broadcast feeds are unknown and thereby, we do not know the values for $H$. Therefore, we tackle this issue using an off-the-shelf homography-estimator \cite{jason-homography} pre-trained on an ice hockey top-view rink template, to map each broadcast frame  onto the overhead view and obtains its respective homographic projection.

\subsection{Temporal Graph Association}

To facilitate association of players across frames, we design a simple temporal graph network, to correlate player features. Inspired by \cite{ReST}, we iteratively correlate the learned graph $G_{t-1}^T$ at time $t-1$ with the next graph $G_{t}$ at time $t$ to form a new graph $G_{t}^T$. Assuming this as the $n^{th}$ iteration, each node $v_{t-1} \in G_{t-1}^T$ contains aggregated embeddings from $n-1$ iterations and is connected with the new nodes $v_{t} \in G_{t}$ to form the new temporal graph $G_{t}^T$ at time $t$ (Ref algorithm \ref{inference-algorithm}). The edge set thus created can be denoted as: 

$$e_{t-1, t} =  
\begin{cases}
    1, \text{if } v_i \times v_j \forall \ v_i \in G_{t-1}, v_j \in G_t\\
    0, \text{otherwise}
\end{cases}
$$

\vspace{3pt}

This aggregation of uniterated graphs with learned graphs helps propagate learned features and assign consistent identities for the same player (Ref Fig. \ref{proposed-workflow})

\begin{algorithm}
\caption{Model Inference}
\label{inference-algorithm}
\begin{algorithmic}[1]
    \State Given g.t $y_{e_{ij}}$, players $P_k \in G_{t-1} \oplus G_t$, $i=0$ \& $w=2$;
    \For{$t=0$ to $(T-1)$} 
        \State Initialize $h^{0}_{v_i}, h^{0}_{v_j}$ and $h^0_{e_{ij}}$ for $G_{t-1} \oplus G_t$
        \If{$ i > 0$}
            \State Replace $G_{t-1}$ with $G^{learned}_t$
        \EndIf
        \newline
        \State $h^{0}_{v_i},h^{0}_{v_j} = f^{FE}_v(h^{0}_{v_i},h^{0}_{v_j})$ 
        \newline
        \State $h^{0}_{e_{ij}} = \it{f}_e^{FE}(h^{0}_{e_{ij}})$ 
        \newline
        \While{$(l \geq L)$}
            \begin{align*} 
            \quad h^{L}_{v_i},h^{L}_{v_j},h^{L}_{e{_{ij}}} = 
            \text{MPN}(G_{t-1} \oplus G_{t}, h^{l-1}_{v_i}, h^{l-1}_{v_j})
            \end{align*}
            \quad \quad \ \ $\hat{y}_{{e}_{ij}} =f^{cls}(h_{e_{ij}}^{L})$ 
        \EndWhile
        \newline
        \State$G_t^{learned} = \text{Post-Processing}(G_{t-1} \oplus G_{t},\hat{y}_{{e}_{ij}}^{L} )$\\
        \quad \ $\text{i++;}$
    \EndFor

\end{algorithmic}
\end{algorithm}

\subsection{Message Passing Network}

We adopt the Message Passing Network (MPN) structure as introduced by \cite{mpn-quantumchem} to propagate the node and edge information across the graph $G$. Message passing intuitively helps the graphs learn their neighboring features; each edge learns about the projection and appearance feature of its neighboring nodes and each node learns about the geometric features of its neighboring edges. To begin with, we initialize the node embeddings and edge embeddings as: 
\begin{equation}
    \label{node-embedding}
    h_{v_i}^{(0)} = f_v^{FE}([ r^{id}_{v_{_i}}, p_{v_{_i}}])
\end{equation}
\begin{equation}
    \label{edge-embedding}
    h_{e_{ij}}^{(0)} = f_e^{FE}([ \Delta r^{id}_{e_{_{ij}}}, \Delta p_{e_{_{ij}}}])
\end{equation}

Note that we add $p$ and $\Delta p$ to both the node and edge features respectively to propagate homographic (positional) information throughout the graph. Given these initial embeddings, as standardized by well-established methods \cite{vatsevaluation, motneuralsolver, ReST, mpn, graph-mcmot}, we perform $L$ iterations of edge updates and node updates as two separate steps. 

\vspace{2pt}
\noindent \textbf{Edge Update}
We utilize a learnable multi-layer perceptron (MLP) to perform edge encoding for $l=\{1, \ldots ,L \}$ steps using the source and destination nodes connected by the edge: 

\begin{equation}
\label{edgeupdate}
(v \rightarrow e) \ \ \ h_{e_{ij}}^{(l)} = f_e^{ME}([h_{v_{i}}^{(l-1)}, h_{v_{j}}^{(l-1)}, h_{e_{ij}}^{(l-1)}])   
\end{equation}

\vspace{3pt}

where, $f_e^{ME}$ is the edge encoder. This leads to the sharing of appearance and projection embeddings from the neighboring nodes $h_{v_{i}}, h_{v_{j}}$ to its connecting edge $h_{e_{ij}}$. 

\vspace{3pt}
\noindent \textbf{Node Update}
Similar to Eq. \eqref{edgeupdate}, we utilize a learnable MLP to perform node update for $l=\{1, \ldots ,L \}$ steps, using the aggregated messages coming from its neighboring nodes: 

\begin{equation}
(e \rightarrow v) \ \ \ h_{v_{i}}^{(l)} = 
\sum_{j \in \mathcal{N}(v_i)}
f_v^{ME}([h_{v_{j}}^{(l-1)}, h_{e_{ij}}^{(l)}])   
\end{equation}

\vspace{3pt}
where, $f_v^{ME}$ is the node encoder and $\mathcal{N}(v_i)$ denotes the neighboring nodes of $v_i$. Note that $f_v^{ME}$ and $f_e^{ME}$ are two separate networks with different dimensions, but share the same MLP architecture (Ref. Table \ref{architecture-mpn})

In both the updates, the MLP encodes all the information into a higher-dimensional feature space. The message passing step $L$ is akin to the receptive field found in convolutional neural networks \cite{alexnet} and a higher value of $L$ corresponds to farther propagation of information in the graph, but at the cost of computation.

\begin{table}[]
\begin{tabular}{|p{1.5cm}||p{2.3cm}|p{1.5cm}|p{1.5cm}|  }
\hline
\quad Network & \quad \quad \quad Layer & \quad \ Input & \quad \ Output \\
\hline
\thickhline
\makecell{\multirow{2}{*}{$\it{f}^{FE}_{v}(\cdot)$}} & \makecell{\multirow{2}{*}{FC+GELU}} & \makecell{515} & \makecell{128} \\
\cline{3-4} & & \makecell{4} & \makecell{32} \\ 
\thickhline
\makecell{\multirow{2}{*}{$\it{f}^{FE}_{e}(\cdot)$ }} & \makecell{\multirow{2}{*}{FC+GELU}} & \makecell{4} & \makecell{8} \\
\cline{3-4} & & \makecell{8} & \makecell{6} \\ 
\thickhline
\makecell{\multirow{2}{*}{$\it{f}^{ME}_{v}(\cdot)$}} & \makecell{\multirow{2}{*}{FC+GELU}} & \makecell{38} & \makecell{64} \\
\cline{3-4} & & \makecell{64} & \makecell{32} \\ 
\thickhline
\makecell{\multirow{2}{*}{$\it{f}^{ME}_{e}(\cdot)$}} & \makecell{\multirow{2}{*}{FC+GELU}} & \makecell{70} & \makecell{32} \\
\cline{3-4} & & \makecell{32} & \makecell{6} \\ 
\thickhline
\makecell{\multirow{2}{*}{$\it{f}^{cls}(\cdot)$}} & \makecell{\\FC+GELU} & \makecell{6} & \makecell{4} \\
\cline{2-4} & \makecell{\\FC+Sigmoid} & \makecell{4} & \makecell{1} \\ 
\thickhline
\end{tabular}
    \caption{Details of each mlp encoder. \\ fc - fully connected, gelu activation \cite{gelu}}
    \label{architecture-mpn}
\end{table}

\vspace{3pt}
\noindent \textbf{Classification}
We propose to learn the association between nodes as a \textit{link prediction task} by framing player tracking as a graph partition problem. That is, after $L$ iterations, we perform binary classification to predict the edge probabilities $\hat y_{e_{ij}}$ connecting nodes $v_i$ and $v_j$, as: 

\begin{equation}
    \label{classifier}
    \hat y_{e_{ij}} = f^{cls}(h_{e_{ij}}) 
\end{equation}
where $f^{cls}$ is a learnable MLP with a sigmoid final layer to output probabilities. During inference, the edges with low confidence scores (weak connections) are removed. During training, the binary ground truth labels $y_{e_{ij}}$ and their corresponding predictions $\hat y_{e_{ij}}$ are compared to find the sigmoid focal loss \cite{sigmoid-focal-loss}.

\subsection{Post-Processing}
We adopt a post-processing step during inference to prune and resolve violations in our final graphs, and assign consistent tracklet IDs. 

\noindent \textbf{Pruning.} This is the first step in refining the predicted edge confidence scores $\hat y_{e_{ij}}$ by our classifier. We define a confidence threshold hyperparameter  $\xi$, where: 

\begin{equation}
    \label{pruning-equation}
    \text {Pruned}(\hat y_{e_{ij}})  =  
    \begin{cases}
        \hat y_{e_{ij}}, \ \ \text{if} \ \hat y_{e_{ij}} > \xi \\
        0, \ \ \text {otherwise}
    \end{cases}
\end{equation}

This helps eliminate the low-confidence edges and retain only the most strongest correspondences. For our experiments, we've used $\xi = 0.9$ to only retain the most strongest edges. Alongside, similar to \cite{ReST}, we remove many-to-one edge violations, wherein, there can only exist at most one connection between two nodes within the connected graphs. This ensures that we represent each player identity through a unique connected component, since no two different players can share the same identity at the same time. 

\noindent \textbf{Assigning IDs.}
The final graph contains unique connected components where each component represents one unique player tracklet. As we iterate temporally, we assign a new identity when there exists no previous connected components for a node or propagate the same identity otherwise. 

\begin{equation}
    \label{assigning-id}
    \text{ID}(h_{v_{j}}) = 
    \begin{cases}
        \text {SameID}, \ \text{if} \ h_{v_{j}} = h_{v_{i}}; \forall j \in G_t, i \in G_{t-1} \\
        \text {NewID}, \ \text{otherwise}
    \end{cases}
\end{equation}

\section{Implementation Details}
This section contains details about datasets used, training scheme, evaluation metrics and the final results. 

\subsection{Datasets} We experiment our methodology on the two available ice hockey tracking datasets - first, similar to the current SOTA benchmark \cite{vatsevaluation}, we train and test on the broadcast hockey dataset for a fair comparison; second, we evaluate on the public VIP Hockey Tracking Dataset (VIP-HTD) \cite{vip-htd} to demonstrate the generalization of our method. Both datasets contain side-of-the-rink broadcast videos with occlusions, blurs, and challenging player movements. 

\vspace{3pt}
\noindent \textbf{Broadcast Tracking Dataset \cite{vatsevaluation}} This dataset contains 84 broadcast clips sampled from 25 NHL games, with an average duration of $\sim$36 seconds per clip. The dataset has a $1280\times720p$ resolution (Standard Definition) at a frame rate of 30fps, with a train:validation:test ratio of 58:13:13. We follow the same training and testing scheme as \cite{vatsevaluation} for equal comparison, and show superior results with our method (Ref Table \ref{eval-broadcasthockey})

\vspace{3pt}
\noindent \textbf{VIP-HTD \cite{vip-htd}} This \emph{public} dataset contains 22 broadcast hockey clips sampled from 8 NHL games, with both 30 \& 60Hz frame rates, recorded at $1280\times720p$ resolution. We perform cross-dataset validation (trained on broadcast dataset; tested on VIP-HTD) on all the 7 test clips in this dataset to showcase the generalizability of our method for any given broadcast hockey feed (Ref Table \ref{evaluation-vip-htd})

\subsection{Training Details.} Given the player bounding boxes and tracklet IDs, we find the footpoint projection coordinates using an off-the-shelf homography model \cite{jason-homography} trained specifically for NHL ice-hockey rinks. This model is currently the SOTA for hockey and helps predict highly accurate overhead projections. Next, we exploit the OSNet architecture \cite{omni-reid} as our ReID network for player feature extraction (Eq. \eqref{reid-equation}), pre-trained on the ImageNet dataset \cite{imagenet}. We encode the 512-D ReID feature vectors along with the 3-D (homogenous coordinates) homography features into 32-D node embeddings (Eq. \eqref{node-embedding}), and the 4-D edge features (Eq. \eqref{edge-embedding}) into 6-D edge embeddings. We run the MPN for L = 6 iterations and pass the final graph output into our binary classifier for predicting edge probabilities. 

\begin{algorithm}
\caption{Model Training}
\begin{algorithmic}[1]
    \State Given g.t $y_{e_{ij}}$, players $P_k \in G_{t-1} \oplus G_t$ \& $w=2$;
    \For{$t=1$ to $(T-1)$} 
        \State Initialize $h^{0}_{v_i},h^{0}_{v_j}$ and $h^0_{e_{ij}}$ for $G_{t-1} \oplus G_t$
        \newline 
        \State $(h^{0}_{v_i},h^{0}_{v_j}) = f^{FE}_v(h^{0}_{v_i},h^{0}_{v_j})$  
        \newline 
        \State $h^{0}_{e_{ij}} = \it{f}_e^{FE}(h^{0}_{e_{ij}})$ 
        \newline 
        \While{\it{l} $\leq$ L}:
            \begin{align*} 
            h^{L}_{v_i},h^{L}_{v_j},h^{L}_{e{_{ij}}} = 
            \text{MPN}(G_{t-1} \oplus G_{t}, h^{l-1}_{v_i}, h^{l-1}_{v_j})
            \end{align*}
            \ \ \ \ \ \ $\hat{y}_{{e}_{ij}} =f^{cls}(h_{e_{ij}}^{L})$ 

        \EndWhile
        \State graph\_loss = sigmoid\_focal\_loss$( \hat{y}_{e_{ij}}, y_{e_{ij}} ) $
       
        \State graph\_loss.backward() \Comment{Backpropagation}
        \State optimizer.step() \Comment{Update parameters} 
    \EndFor


\end{algorithmic}
\end{algorithm}

During training, we utilize the ground-truth player annotations and calculate the prediction losses using Focal Loss \cite{sigmoid-focal-loss}

\begin{equation}
    \label{focal-loss}
    \text{Focal Loss} = 
    \sum_l^L \sum_{e_{ij} \in G_{t-1} \cup G_t}
    \mathcal{L}(\hat y^l_{e_{ij}}, y_{e_{ij}} )
\end{equation}

where, $\hat y^l_{e_{ij}}$ is the edge prediction at iteration $l$ and $y_{e_{ij}}$ is the ground-truth indicator function: 
\begin{equation}
    \label{gt-indicator}
    \mathds{1}(y_{e_{ij}}) = 
    \begin{cases}
        1, \ \text{if} \  v_{i} = v_{j} \\
        0, \  \text{otherwise}
    \end{cases}
\end{equation}

that is, for every player match, $y_{e_{ij}}$ = 1 else 0. We use Adam Optimizer \cite{adam} without weight decays to update our model parameters. The learning rate (LR) is initialized at 0.01, with a gradual warmup for the first 10 epochs and a cosine annealing schedule (min. LR = 0.001) thereafter. We trained our model for 30 epochs with a batch size of 16 on a single NVIDIA GeForce RTX 4090 GPU with 24 GB RAM, 2.5GHz clock speed, and performed validation after every 2 training epochs. 

\begin{table*}
    \centering
    \begin{tabular}{c|c|c|c|c|c}
        \hline
         Method & MOTA \% $\uparrow$ & FP $\downarrow$ & FN $\downarrow$ & IDsw $\downarrow$ & IDF1 $\uparrow$ \% \\
        \hline
        SORT   \cite{SORT}      &  92.4 & 2403 &  5826 & 673 & 53.7 \\ 
        DeepSORT \cite{deepSORT}   & 94.2  &  1881 &  4334 & 528 & 59.3\\ 
        FairMOT  \cite{fairmot}   &  91.9 &  \textbf{1179} &  7568 & 768 & 61.5 \\ 
        Tracktor  \cite{tracktor}  & 94.4  & 1706 &  \textbf{4216} & 687 & 56.5 \\
        Hockey MOT \cite{vatsevaluation}  & 94.5  &  1653 & 4394 & \textbf{431} & 62.9\\
        Hockey MOT\textsuperscript{\dag} \cite{vatsevaluation}  & - &  - & - & 1056 & 71.8\\
        \hline
        \textbf{Ours} & \textbf{95.4} & 1924 & 4323 & 453 & \textbf{71.3}\\
        \textbf{Ours\textsuperscript{\dag}} & - & - & - & \textbf{151} & \textbf{95.1} \\
        \hline
    \end{tabular}
    \caption{Evaluation results on the broadcast ice hockey dataset. \textsuperscript{\dag} indicates ground-truth annotations used for evaluation}
    \label{eval-broadcasthockey}
\end{table*}

\subsection{Evaluation Metrics} \label{Evaluation}

Most common evaluation metrics used in popular SOTA tracking methods are the Multi-Object Tracking Accuracy (MOTA)\cite{clearmotmetrics} and IDF1 score \cite{idf1score}. With respect to our problem context, they can be defined as: 

\vspace{3pt}
\noindent \textbf{MOT Accuracy:} It is calculated as the complement of three distinct errors - 
\begin{itemize}
    \item False Positives (FP): No. of false players detected; 
    \item False Negatives (FN): No. of true players missed;
    \item Identity Switches (IDsw): No. of identity swaps/re-initializations made for players within the field-of-view. 
\end{itemize}

\begin{equation}
    \label{motaccuracy-eqn}
    \text{MOTA} = 1 - \frac{\sum_{t} \text{FN}_t + \text{FP}_t + \text{IDsw}_t)}{\sum_{t} \text{GT}_t}
\end{equation}
where, $\text{GT}_t$ denotes the ground-truth annotations. 

\vspace{3pt}
\noindent \textbf{IDF1 Score:} It is defined as the ratio of correctly identified
players over the average number of ground-truth and computed identities: 

\begin{equation}
    \label{idf1-eqn}
   \text{IDF1} = 2 \times \frac{{\text{TP}_{id}}}{(2 \times {\text{TP}_{id})} + {\text{FP}_{id}} + {\text{FN}_{id}}}
\end{equation}

where, $\text{TP}_{id}, \text{FP}_{id}, \text{FN}_{id}$ are True Positive, False Positive and False Negative player identities respectively. Alternatively, IDF1 score can also be defined as the harmonic mean of ID Precision and ID Recall.


\begin{table*}
    \centering
    
    \begin{tabular}{c|c|c|c|c}
        \hline
        NHL teams (clips) & No.of frames & FPS (Hz) & IDsw $\downarrow$ & IDF1 $\uparrow$ \\
        \hline
         CAR vs. BOS & 2606 & 60 & 115&85.4 \\ 
         CAR vs. NYR & 2137 & 60 & 71 & 81.1 \\
         CGY vs. DAL & 1330 & 60 & 41&77.4 \\
         CHI vs. TOR & 1618 & 30& 152&60.3 \\
         CNTRL vs. PAC & 1671 & 30 & 136& 71.7\\
         PIT vs. SJ & 1391 & 30 & 164&70.1 \\
         STL vs. SJ & 2258 & 60 & 108& 86.2\\
         \hline
         Hockey MOT  &  & & 787& 80.2 \\
         \hline
         CAR vs. BOS & 2606  & 60 & 11 & 89.3\\ 
         CAR vs. NYR & 2137 & 60 & 7  & 93.1 \\
         CGY vs. DAL & 1330 & 60 & 6 & 92.4\\
         CHI vs. TOR & 1618 & 30 & 6 & 90.9\\
         CNTRL vs. PAC & 1671 & 30 & 13 & 90.5 \\
         PIT vs. SJ & 1391 & 30 & 7 & 97.4 \\
         STL vs. SJ & 2258 & 60 & 7 & 96.3 \\
        \hline
         \textbf{Ours}  & 13,011 & - & \textbf{60} & \textbf{92.84}\\
        \hline 
    \end{tabular}
    \caption{cross-dataset validation on the public vip-hockey tracking dataset}
    \label{evaluation-vip-htd}
\end{table*}

The FPs and FNs used to calculate MOTA relies solely on the detector's quality. Even if the tracker consistently associates players, the MOTA will be skewed if there exists high FP and FN detections, as they have twice the weightage of IDsw in MOTA. Since this doesn't give a clear picture of the tracker's association capabilities, we focus only on the IDsw score and the IDF1 score as the \emph{key} metrics in player tracking. These metrics are especially relevant in ice hockey, since they measure how consistently a player is tracked with respect to his original identity. Therefore, our primary objective is to have $\downarrow$ IDsw and $\uparrow$ IDF1 score for consistent player tracking. Our preferred evaluations are directly based on ground-truth annotations; but, to be consistent with \cite{vatsevaluation} we use similar detection outputs to show our results.

\subsection{Results}

We report the results of our model's performance on the test-sets of the broadcast ice hockey dataset \cite{vatsevaluation} and the public VIP-HTDataset \cite{vip-htd}. It is to be noted that we reproduce the results of our benchmark \cite{vatsevaluation}, under similar hardware and testing conditions as our own method's evaluations, to avoid any discrepancies. From Table \ref{eval-broadcasthockey} during ground-truth evaluations, our model outperforms the SOTA model by a large 23.3\% $\uparrow$ in \emph{IDF1} score, and ~10$\times \downarrow$ in IDsw. This is due to the ability of our model to handle heavy occlusions and blurs prevalent in these videos. We see similar trends with the F-RCNN \cite{faster-rcnn} detection inputs, where our model surpasses all methods by a 8.4\%$\uparrow$ in \emph{IDF1}. Our MOT Accuracy is still higher than all other methods, despite incurring more FPs and FNs due to the ability of our tracker to recover robustly from mistakes made by the detector. We show qualitative results for all 7 videos below the reference section. In Table \ref{evaluation-vip-htd}, we cross-validate our model on the VIP-HTDataset \cite{vip-htd} showing a clear superiority in both \emph{IDsw} and \emph{IDF1} metrics, compared to \cite{vatsevaluation}. This asserts two important things: (i) our model generalizes well to unseen hockey feeds, despite of varying environmental conditions; (ii) Our model's performance isn't affected by the $\uparrow$ in frame rate.   

\section{Conclusion}
We present a novel approach based on the combination of graphical neural networks and homography to effectively track ice hockey players in broadcast feeds. We project player footpoints to an overhead rink template to maintain consistent positional cues, especially during occlusions and blurry situations. This provides a pseudo 'top-view' effect to disentangle overlapping players and maintain their trajectories. Message passing network (MPN) is used to aggregate player features and model their temporal relationships, followed by a classifier to predict player association probabilities. We achieve as significant $\uparrow$ in $IDF1$ and $\downarrow$ in $IDsw$, when compared to both the current SOTA benchmark and a public tracking dataset. We believe that our work can also benefit various other sports in the future.

\section*{Acknowledgment}

Thanks to Jerrin Bright for helping with the figures and formatting. This work was supported by Stathletes through the Mitacs Accelerate Program and the Natural Sciences and Engineering Research Council of Canada (NSERC).




%


\newpage

\bibliographystyle{IEEEtran}
\bibliography{crv}

\begin{thebibliography}{10}
\providecommand{\url}[1]{#1}
\csname url@samestyle\endcsname
\providecommand{\newblock}{\relax}
\providecommand{\bibinfo}[2]{#2}
\providecommand{\BIBentrySTDinterwordspacing}{\spaceskip=0pt\relax}
\providecommand{\BIBentryALTinterwordstretchfactor}{4}
\providecommand{\BIBentryALTinterwordspacing}{\spaceskip=\fontdimen2\font plus
\BIBentryALTinterwordstretchfactor\fontdimen3\font minus \fontdimen4\font\relax}
\providecommand{\BIBforeignlanguage}[2]{{%
\expandafter\ifx\csname l@#1\endcsname\relax
\typeout{** WARNING: IEEEtran.bst: No hyphenation pattern has been}%
\typeout{** loaded for the language `#1'. Using the pattern for}%
\typeout{** the default language instead.}%
\else
\language=\csname l@#1\endcsname
\fi
#2}}
\providecommand{\BIBdecl}{\relax}
\BIBdecl

\bibitem{mot2015}
L.~Leal-Taixé, A.~Milan, I.~Reid, S.~Roth, and K.~Schindler, ``Motchallenge 2015: Towards a benchmark for multi-target tracking,'' 2015.

\bibitem{mot16}
A.~Milan, L.~Leal-Taixe, I.~Reid, S.~Roth, and K.~Schindler, ``Mot16: A benchmark for multi-object tracking,'' 2016.

\bibitem{mot20}
P.~Dendorfer, H.~Rezatofighi, A.~Milan, J.~Shi, D.~Cremers, I.~Reid, S.~Roth, K.~Schindler, and L.~Leal-Taixé, ``Mot20: A benchmark for multi object tracking in crowded scenes,'' 2020.

\bibitem{dancetrack}
P.~Sun, J.~Cao, Y.~Jiang, Z.~Yuan, S.~Bai, K.~Kitani, and P.~Luo, ``Dancetrack: Multi-object tracking in uniform appearance and diverse motion,'' in \emph{Proceedings of the IEEE/CVF Conference on Computer Vision and Pattern Recognition (CVPR)}, 2022.

\bibitem{kitti-autonomous-vehicles}
A.~Geiger, P.~Lenz, and R.~Urtasun, ``Are we ready for autonomous driving? the kitti vision benchmark suite,'' in \emph{Conference on Computer Vision and Pattern Recognition (CVPR)}, 2012.

\bibitem{alexnet}
A.~Krizhevsky, I.~Sutskever, and G.~Hinton, ``Imagenet classification with deep convolutional neural networks,'' \emph{Neural Information Processing Systems}, vol.~25, 01 2012.

\bibitem{soccertrackingsurvey}
M.~Manafifard, H.~Ebadi, and H.~A. Moghaddam, ``A survey on player tracking in soccer videos,'' \emph{Computer Vision and Image Understanding}, vol. 159, pp. 19--46, 2017.

\bibitem{handball-tracking}
M.~Buric, M.~Ivasic-Kos, and M.~Pobar, ``Player tracking in sports videos,'' in \emph{2019 IEEE International Conference on Cloud Computing Technology and Science (CloudCom)}, 2019, pp. 334--340.

\bibitem{basketball1}
M.-C. Hu, M.-H. Chang, J.-L. Wu, and L.~Chi, ``Robust camera calibration and player tracking in broadcast basketball video,'' \emph{IEEE Transactions on Multimedia}, vol.~13, no.~2, pp. 266--279, 2011.

\bibitem{basketball2}
W.-L. Lu, J.-A. Ting, J.~J. Little, and K.~P. Murphy, ``Learning to track and identify players from broadcast sports videos,'' \emph{IEEE transactions on pattern analysis and machine intelligence}, vol.~35, no.~7, pp. 1704--1716, 2013.

\bibitem{basketball3}
H.-T. Chen, C.-L. Chou, T.-S. Fu, S.-Y. Lee, and B.-S.~P. Lin, ``Recognizing tactic patterns in broadcast basketball video using player trajectory,'' \emph{Journal of Visual Communication and Image Representation}, vol.~23, no.~6, pp. 932--947, 2012.

\bibitem{volleyball1}
M.~Takahashi, K.~Ikeya, M.~Kano, H.~Ookubo, and T.~Mishina, ``Robust volleyball tracking system using multi-view cameras,'' in \emph{2016 23rd International Conference on Pattern Recognition (ICPR)}.\hskip 1em plus 0.5em minus 0.4em\relax IEEE, 2016, pp. 2740--2745.

\bibitem{sportsmot}
Y.~Cui, C.~Zeng, X.~Zhao, Y.~Yang, G.~Wu, and L.~Wang, ``Sportsmot: A large multi-object tracking dataset in multiple sports scenes,'' 2023.

\bibitem{soccernet}
A.~Cioppa, S.~Giancola, A.~Deliege, L.~Kang, X.~Zhou, Z.~Cheng, B.~Ghanem, and M.~Van~Droogenbroeck, ``Soccernet-tracking: Multiple object tracking dataset and benchmark in soccer videos,'' in \emph{Proceedings of the IEEE/CVF Conference on Computer Vision and Pattern Recognition}, 2022, pp. 3491--3502.

\bibitem{okuma}
K.~Okuma, A.~Taleghani, D.~Freitas, J.~Little, and D.~Lowe, ``A boosted particle filter: Multitarget detection and tracking,'' vol. 3021, 05 2004.

\bibitem{adaboost}
P.~Viola and M.~Jones, ``Rapid object detection using a boosted cascade of simple features,'' in \emph{Proceedings of the 2001 IEEE Computer Society Conference on Computer Vision and Pattern Recognition. CVPR 2001}, vol.~1, 2001, pp. I--I.

\bibitem{mixture-particle-filter}
Vermaak, Doucet, and Perez, ``Maintaining multimodality through mixture tracking,'' in \emph{Proceedings Ninth IEEE International Conference on Computer Vision}, 2003, pp. 1110--1116 vol.2.

\bibitem{cai}
Y.~Cai, N.~Freitas, and J.~Little, ``Robust visual tracking for multiple targets,'' 05 2006, pp. 107--118.

\bibitem{vatsevaluation}
K.~Vats, M.~Fani, D.~A. Clausi, and J.~S. Zelek, ``Evaluating deep tracking models for player tracking in broadcast ice hockey video,'' 2022.

\bibitem{motneuralsolver}
G.~Brasó and L.~Leal-Taixé, ``Learning a neural solver for multiple object tracking,'' 2020.

\bibitem{vatstrackingandid}
K.~Vats, P.~Walters, M.~Fani, D.~A. Clausi, and J.~Zelek, ``Player tracking and identification in ice hockey,'' 2021.

\bibitem{jason-homography}
J.~C. Shang, Y.~Chen, M.~J. Shafiee, and D.~A. Clausi, ``Rink-agnostic hockey rink registration,'' 2023.

\bibitem{mpn}
P.~W. Battaglia, J.~B. Hamrick, V.~Bapst, A.~Sanchez-Gonzalez, V.~Zambaldi, M.~Malinowski, A.~Tacchetti, D.~Raposo, A.~Santoro, R.~Faulkner, C.~Gulcehre, F.~Song, A.~Ballard, J.~Gilmer, G.~Dahl, A.~Vaswani, K.~Allen, C.~Nash, V.~Langston, C.~Dyer, N.~Heess, D.~Wierstra, P.~Kohli, M.~Botvinick, O.~Vinyals, Y.~Li, and R.~Pascanu, ``Relational inductive biases, deep learning, and graph networks,'' 2018.

\bibitem{mpn-quantumchem}
J.~Gilmer, S.~S. Schoenholz, P.~F. Riley, O.~Vinyals, and G.~E. Dahl, ``Neural message passing for quantum chemistry,'' 2017.

\bibitem{faster-rcnn}
\BIBentryALTinterwordspacing
S.~Ren, K.~He, R.~B. Girshick, and J.~Sun, ``Faster {R-CNN:} towards real-time object detection with region proposal networks,'' \emph{CoRR}, vol. abs/1506.01497, 2015. [Online]. Available: \url{http://arxiv.org/abs/1506.01497}
\BIBentrySTDinterwordspacing

\bibitem{smiletrack}
Y.-H. Wang, J.-W. Hsieh, P.-Y. Chen, M.-C. Chang, H.~H. So, and X.~Li, ``Smiletrack: Similarity learning for occlusion-aware multiple object tracking,'' 2024.

\bibitem{ucmctrack}
K.~Yi, K.~Luo, X.~Luo, J.~Huang, H.~Wu, R.~Hu, and W.~Hao, ``Ucmctrack: Multi-object tracking with uniform camera motion compensation,'' 2024.

\bibitem{bytetrack}
Y.~Zhang, P.~Sun, Y.~Jiang, D.~Yu, F.~Weng, Z.~Yuan, P.~Luo, W.~Liu, and X.~Wang, ``Bytetrack: Multi-object tracking by associating every detection box,'' 2022.

\bibitem{tracktor}
\BIBentryALTinterwordspacing
P.~Bergmann, T.~Meinhardt, and L.~Leal-Taixe, ``Tracking without bells and whistles,'' in \emph{2019 {IEEE}/{CVF} International Conference on Computer Vision ({ICCV})}.\hskip 1em plus 0.5em minus 0.4em\relax {IEEE}, oct 2019. [Online]. Available: \url{https://doi.org/10.1109\%2Ficcv.2019.00103}
\BIBentrySTDinterwordspacing

\bibitem{transmot}
P.~Chu, J.~Wang, Q.~You, H.~Ling, and Z.~Liu, ``Transmot: Spatial-temporal graph transformer for multiple object tracking,'' \emph{2023 IEEE/CVF Winter Conference on Applications of Computer Vision (WACV)}, pp. 4859--4869, 2021.

\bibitem{SORT}
\BIBentryALTinterwordspacing
A.~Bewley, Z.~Ge, L.~Ott, F.~Ramos, and B.~Upcroft, ``Simple online and realtime tracking,'' \emph{CoRR}, vol. abs/1602.00763, 2016. [Online]. Available: \url{http://arxiv.org/abs/1602.00763}
\BIBentrySTDinterwordspacing

\bibitem{deepSORT}
\BIBentryALTinterwordspacing
N.~Wojke, A.~Bewley, and D.~Paulus, ``Simple online and realtime tracking with a deep association metric,'' \emph{CoRR}, vol. abs/1703.07402, 2017. [Online]. Available: \url{http://arxiv.org/abs/1703.07402}
\BIBentrySTDinterwordspacing

\bibitem{fairmot}
\BIBentryALTinterwordspacing
Y.~Zhang, C.~Wang, X.~Wang, W.~Zeng, and W.~Liu, ``Fairmot: On the fairness of detection and re-identification in multiple object tracking,'' \emph{International Journal of Computer Vision}, vol. 129, no.~11, 2021. [Online]. Available: \url{http://dx.doi.org/10.1007/s11263-021-01513-4}
\BIBentrySTDinterwordspacing

\bibitem{kalmanfilters}
R.~E. Kalman, ``A new approach to linear filtering and prediction problems,'' 1960.

\bibitem{hungarianmethod}
H.~Kuhn, ``The hungarian method for the assignment problem,'' \emph{Naval Research Logistic Quarterly}, vol.~2, 05 2012.

\bibitem{centernet}
\BIBentryALTinterwordspacing
X.~Zhou, D.~Wang, and P.~Kr{\"{a}}henb{\"{u}}hl, ``Objects as points,'' \emph{CoRR}, vol. abs/1904.07850, 2019. [Online]. Available: \url{http://arxiv.org/abs/1904.07850}
\BIBentrySTDinterwordspacing

\bibitem{sparsetrack}
Z.~Liu, X.~Wang, C.~Wang, W.~Liu, and X.~Bai, ``Sparsetrack: Multi-object tracking by performing scene decomposition based on pseudo-depth,'' 2023.

\bibitem{iwase-soccer}
S.~Iwase and H.~Saito, ``Parallel tracking of all soccer players by integrating detected positions in multiple view images,'' in \emph{Proceedings of the 17th International Conference on Pattern Recognition, 2004. ICPR 2004.}, vol.~4, 2004, pp. 751--754 Vol.4.

\bibitem{ming-soccer}
M.~Xu, J.~Orwell, and G.~Jones, ``Tracking football players with multiple cameras,'' in \emph{2004 International Conference on Image Processing, 2004. ICIP '04.}, vol.~5, 2004, pp. 2909--2912 Vol. 5.

\bibitem{bayesian-soccer}
P.~Nillius, J.~Sullivan, and S.~Carlsson, ``Multi-target tracking - linking identities using bayesian network inference,'' in \emph{2006 IEEE Computer Society Conference on Computer Vision and Pattern Recognition (CVPR'06)}, vol.~2, 2006, pp. 2187--2194.

\bibitem{initial-graph-2004}
P.~Figueroa, N.~Leite, R.~Barros, I.~Cohen, and G.~Medioni, ``Tracking soccer players using the graph representation,'' in \emph{Proceedings of the 17th International Conference on Pattern Recognition, 2004. ICPR 2004.}, vol.~4, 2004, pp. 787--790 Vol.4.

\bibitem{Yolo}
J.~Redmon, S.~K. Divvala, R.~B. Girshick, and A.~Farhadi, ``You only look once: Unified, real-time object detection,'' \emph{2016 IEEE Conference on Computer Vision and Pattern Recognition (CVPR)}, pp. 779--788, 2015.

\bibitem{basketball-SORT}
\BIBentryALTinterwordspacing
D.~Acuna, ``Towards real-time detection and tracking of basketball players using deep neural networks.'' [Online]. Available: \url{https://api.semanticscholar.org/CorpusID:31248790}
\BIBentrySTDinterwordspacing

\bibitem{theagarajan}
R.~Theagarajan and B.~Bhanu, ``An automated system for generating tactical performance statistics for individual soccer players from videos,'' \emph{IEEE Transactions on Circuits and Systems for Video Technology}, vol.~31, no.~2, pp. 632--646, 2021.

\bibitem{jdegraphnetwork}
Y.~Wang, K.~Kitani, and X.~Weng, ``Joint object detection and multi-object tracking with graph neural networks,'' 2021.

\bibitem{graph-mcmot}
E.~Luna, J.~C. SanMiguel, J.~M. Martínez, and P.~Carballeira, ``Graph neural networks for cross-camera data association,'' 2022.

\bibitem{ReST}
C.-C. Cheng, M.-X. Qiu, C.-K. Chiang, and S.-H. Lai, ``Rest: A reconfigurable spatial-temporal graph model for multi-camera multi-object tracking,'' 2023.

\bibitem{omni-reid}
K.~Zhou, Y.~Yang, A.~Cavallaro, and T.~Xiang, ``Omni-scale feature learning for person re-identification,'' in \emph{ICCV}, 2019.

\bibitem{gelu}
D.~Hendrycks and K.~Gimpel, ``Gaussian error linear units (gelus),'' 2023.

\bibitem{sigmoid-focal-loss}
T.-Y. Lin, P.~Goyal, R.~Girshick, K.~He, and P.~Dollár, ``Focal loss for dense object detection,'' 2018.

\bibitem{vip-htd}
H.~Prakash, Y.~Chen, S.~Rambhatla, D.~Clausi, and J.~Zelek, ``Vip-htd: A public benchmark for multi-player tracking in ice hockey,'' in \emph{Computer Vision and Intelligent Systems}, 2024.

\bibitem{imagenet}
J.~Deng, W.~Dong, R.~Socher, L.-J. Li, K.~Li, and L.~Fei-Fei, ``Imagenet: A large-scale hierarchical image database,'' in \emph{2009 IEEE Conference on Computer Vision and Pattern Recognition}, 2009, pp. 248--255.

\bibitem{adam}
D.~P. Kingma and J.~Ba, ``Adam: A method for stochastic optimization,'' 2017.

\bibitem{clearmotmetrics}
K.~Bernardin and R.~Stiefelhagen, ``Evaluating multiple object tracking performance: The clear mot metrics,'' \emph{EURASIP Journal on Image and Video Processing}, vol. 2008, 01 2008.

\bibitem{idf1score}
E.~Ristani, F.~Solera, R.~S. Zou, R.~Cucchiara, and C.~Tomasi, ``Performance measures and a data set for multi-target, multi-camera tracking,'' 2016.

\end{thebibliography}

\newpage

{
\huge
\textbf{Supplementary}
}

\vspace{5pt}

We present empirical results from experimenting with different network choices and our reasoning behind the proposed framework. This supplementary includes: (i) Effect of Homography; (ii) Effect of message passing steps; (iii) Video-wise quantitative results on the 13-test set videos in broadcast hockey dataset \cite{vatsevaluation}; (iv) Qualitative tracking results on VIP-HTD and broadcast hockey dataset. 

\vspace{3pt}
\noindent \textbf{1. Effect of Homography}

\begin{table}[ht]
    \centering
    \begin{tabular}{c|c|c|c|c}
         \hline 
        Test video \# & Appearance & Homography  & IDsw $\downarrow$ & IDF1 $\uparrow$ \\
        \hline

         & \checkmark & & 31 & 94.8 \\
        1 & \checkmark & \checkmark & \textbf{8} & \textbf{95.7} \\
        \hline
        
         & \checkmark & & 37 & 90.7 \\
        2 & \checkmark & \checkmark & \textbf{27} & \textbf{94.1} \\
        \hline

         & \checkmark & & 19 & 93.5 \\
        3 & \checkmark & \checkmark & \textbf{14} & \textbf{93.9} \\
        \hline

         & \checkmark &  & 13 & 93.8 \\
        4 & \checkmark & \checkmark & \textbf{4} & \textbf{98.2} \\
        \hline

         & \checkmark &  & 53 & 89.2\\
        5 & \checkmark & \checkmark & \textbf{13} & \textbf{97.5} \\
        \hline

         & \checkmark &  & 25 & 90.0 \\
        6 & \checkmark & \checkmark & \textbf{12} & \textbf{94.0} \\
        \hline

         & \checkmark & & 28 & 87.9 \\
        7 & \checkmark & \checkmark & \textbf{11} & \textbf{88.8}\\
        \hline

         & \checkmark & & 35 & 79.0 \\
        8 & \checkmark & \checkmark & \textbf{2} & \textbf{97.9}\\
        \hline

         & \checkmark & & 16 & 92.8 \\
        9 & \checkmark & \checkmark & \textbf{6} & \textbf{97.1}\\
        \hline
        
         & \checkmark & & 25 & 81.5 \\
        10 & \checkmark & \checkmark & \textbf{23} & \textbf{90.8 }\\
        \hline 

         & \checkmark & & 25 & 94.1 \\
        11 & \checkmark & \checkmark & \textbf{4} &\textbf{ 97.6} \\
        \hline

         & \checkmark &  & 23 & 88.2 \\
        12 & \checkmark & \checkmark & \textbf{16} & 92.9 \\
        \hline
        
         & \checkmark &  &\textbf{ 10 }& 94.9 \\
        13 & \checkmark & \checkmark & 11 &\textbf{ 97.7} \\
        \hline
        
    \end{tabular}
    \caption{Effect of homography on IDsw and IDF1 scores for the broadcast ice hockey test-set \cite{vatsevaluation}}
    \label{homo-ablation}
\end{table}

From Table \ref{homo-ablation}, it can be observed that only using appearance features underperforms by a noticeable margin in each listed sequence, while augmenting it with homography cues boosts the IDF1 score and reduces the number of incurred IDsw. This empirically validates our core hypothesis that homography makes tracking better. 


\begin{figure}[H]    
    \centering
    \includegraphics[width=0.65\linewidth]{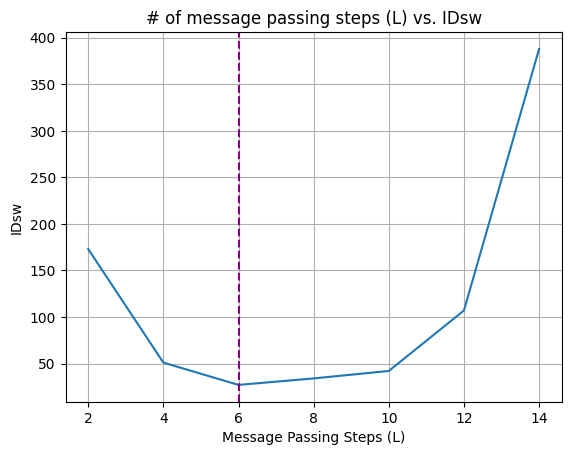}
    \vspace{-10px}
    \caption{No. of message passing steps, $L$ vs. No. of IDsw incurred}
    \label{message-passing-fig}
\end{figure}

\noindent \textbf{2. No. of message passing steps, $L$}

The no. of message passing steps for $L$, analogous to the receptive field of a CNN is best pronounced at $L = 6$ steps in our model. $L<6$ is inadequate to propagate information across neighboring nodes and edges, leading to lesser capability (as evident from the $\uparrow$ IDsw), while $L>6$ leads to possible overfitting on the training data. Thus, $L$ should be treated as an empirical hyperparameter which is best chosen when observed. 

\begin{table}[ht]
    \centering
    \begin{tabular}{c|c|c}
        \hline
         \# of message passing steps & IDsw $\downarrow$ & IDF1 $\uparrow$   \\
         \hline
        
         2 & 173 & 85.5 \\
         4 & 51 & 93.4 \\ 
         \textbf{6} & \textbf{27} & \textbf{94.1} \\
         8 & 34 & 92.2 \\
         10 & 42 & 84.7 \\
         12 & 107 & 67.4 \\
         14 & 388 & 43.3\\
         \hline
    \end{tabular}
    \label{message-passing-ablation}
    \caption{Effect of message passing steps on IDsw and IDF1 scores for a randomly chosen sequence from the broadcast hockey dataset \cite{vatsevaluation}}
\end{table}


\vspace{3pt}
\noindent \textbf{3. Video-wise quantitative results}

Table \ref{eval-13testvids} presents a detailed summary of our test results on the 13 test videos in broadcast ice hockey dataset \cite{vatsevaluation}. Our model achieves SOTA results on all sequences despite occlusions and camera movements. 

\begin{table}[H]
    \centering
    \begin{tabular}{c|c|c|c|c|c}
        \hline
        Video \# & Frames & IDsw \textsuperscript{$\pm$}$\downarrow$ & IDF1 \textsuperscript{$\pm$}$\uparrow$ \% & IDsw $\downarrow$ & IDF1 $\uparrow$ \% \\
        \hline
        1  & 1,303 & 85&  80.7 & 8 & 95.7\\
        2  & 1,210 & 113& 75.4 & 27 & 94.1   \\
        3  & 878 & 114& 77.0 & 14 & 93.9    \\
        4  & 1,033 & 45&  79.7 & 4  & 98.2   \\
        5  & 1,071 & 154&  54.9& 13  & 97.5  \\
        6  & 1,441 & 81&  76.1&  12 &  94.0  \\
        7  & 682 & 58& 63.0 &   11 &   88.8  \\
        8  & 1,056 & 50& 80.7 &  2 &   97.9  \\
        9  & 1,497 & 87& 64.4 &  6 &   97.1 \\
        10  & 890 & 54& 55.9  & 23  &  90.8  \\
        11  & 1,109 & 95& 80.3 &  4 & 97.6  \\
        12  & 1,372 & 56& 72.44 & 16  &  92.9 \\
        13  & 795 & 64& 75.15 &  11  &  97.7 \\
        \hline  
        Overall & 14,337 & 1056 & 71.98 & \textbf{151} & \textbf{95.1} \\
    \hline        
    \end{tabular}
    \caption{Sequence-wise results on the 13 test-set videos from broadcast ice hockey dataset. We evaluate both methods on ground truth annotations.   $\pm$ denotes current sota benchmark results}
    \label{eval-13testvids}
\end{table}

\begin{figure*}
\centering
    \includegraphics[width=\textwidth]{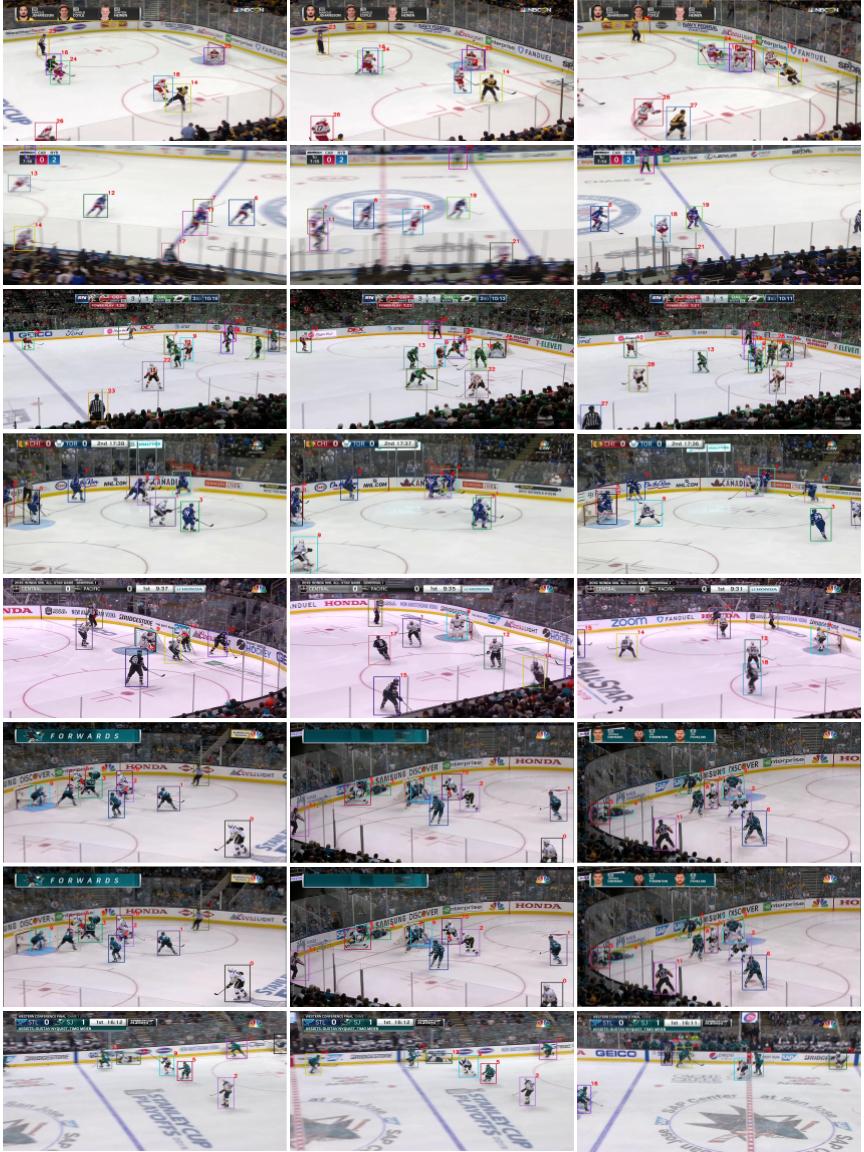}
    \caption{Qualitative results for the VIP-HTD \cite{vip-htd} test-set. Our tracker generalizes well to this unseen dataset, incurring neglible ID switches}
    \label{message-fig}
\end{figure*}

\vspace{3pt}
\noindent \textbf{{4. Qualitative results}}

We show qualitative results for our tracker on the 7 games test-set from VIP-HTD \cite{vip-htd}. It can be observed that our tracker generalizes well to unseen, out-of-distribution (OOD) data incurring less IDsw and excellent IDF1 score (Ref. Table \ref{evaluation-vip-htd}), despite varying lighting conditions, jersey colours, and camera angles. We hope is to further test the generalization capacity of our model across sports in the future.


\end{document}